%% file: neurips_2026.tex
\documentclass{article}

\usepackage[preprint]{neurips_2026}

\input{preamble.tex}

\title{SoccerLens: Grounded Soccer \\  Video Understanding Beyond Accuracy}

\author{%
  Ismael Elsharkawi$^{1}$ \quad 
  Ahmed Sait$^{2}$ \quad 
  Silvio Giancola$^{2}$ \\
  \textbf{Bernard Ghanem}$^{2}$ \quad 
  \textbf{Hossam Sharara}$^{1}$ \quad 
  \textbf{Abdelrahman Eldesokey}$^{2}$ \\
  \\
  $^1$Department of Computer Science and Engineering, The American University in Cairo \\
  $^2$Image And Visual Understanding Lab (IVUL), KAUST \\
  \\
  \texttt{\{ismaelelsharkawi, hossam.sharara\}@aucegypt.edu} \\
  \texttt{\{ahmed.sait, silvio.giancola, bernard.ghanem, abdelrahman.eldesokey\}@kaust.edu.sa}
}

\begin{document}

\maketitle

\begin{figure}[ht]
    \centering
    \includegraphics[width=0.9\linewidth]{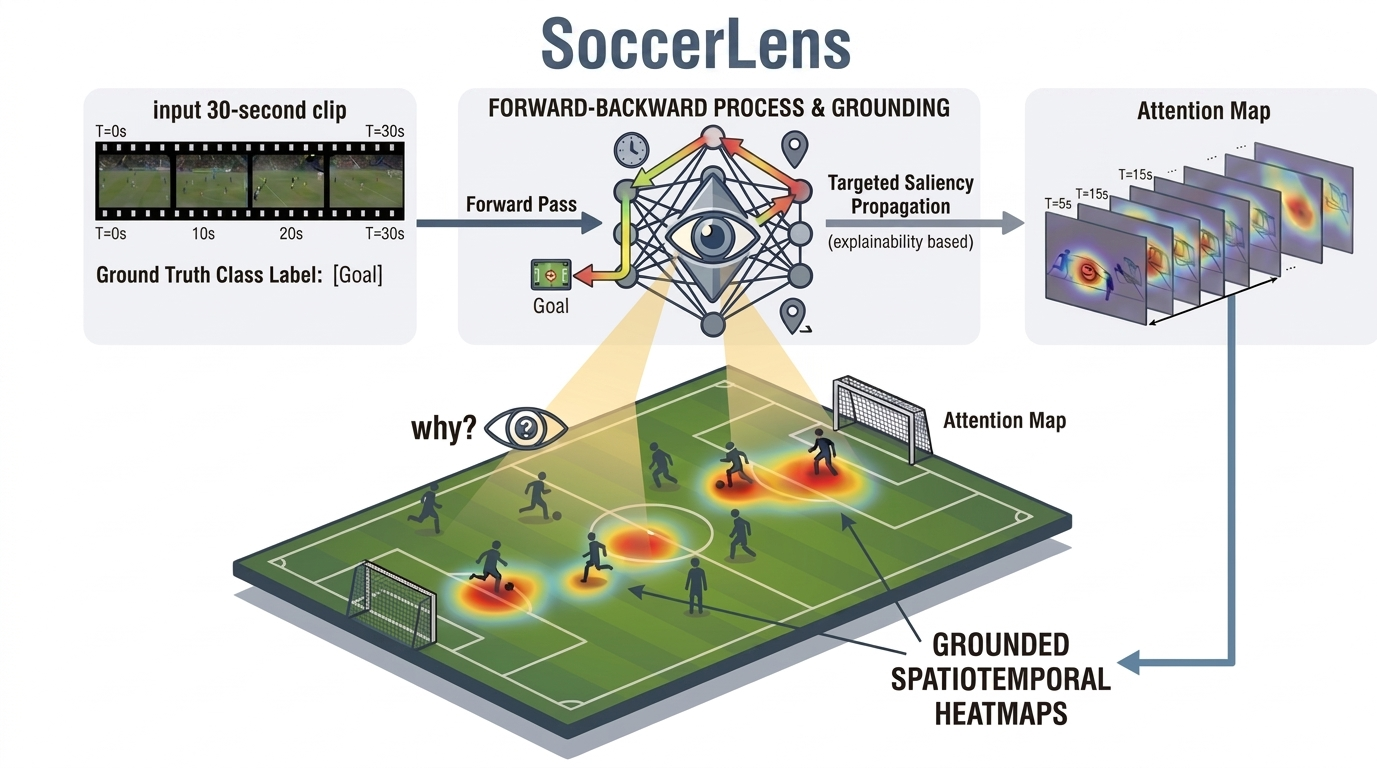}
    \caption{\method introduces a benchmark suite for evaluating visual grounding in soccer VLMs, \ie, whether model predictions are based on semantically relevant regions or spurious cues.}
    \label{fig:teaser}
\end{figure}

\input{sections/0_abstract}

\input{sections/1_intro}
\input{sections/2_related}

\input{sections/3_dataset}

\input{sections/4_prel}
\input{sections/6_exp}

\input{sections/7_discussion}

\input{sections/8_conclusion}

\bibliographystyle{abbrv} 
\bibliography{sn-bibliography}

\input{sections/appendixA}
\input{sections/appendixB}

\end{document}

%% file: preamble.tex
\usepackage[utf8]{inputenc}
\usepackage[T1]{fontenc}

\usepackage[dvipsnames]{xcolor}

\usepackage{graphicx}
\usepackage{multirow}
\usepackage{amsmath,amssymb,amsfonts}
\usepackage{amsthm}
\usepackage{mathrsfs}
\usepackage[title]{appendix}
\usepackage{textcomp}
\usepackage{manyfoot}
\usepackage{booktabs}
\usepackage{algorithm}
\usepackage{algorithmicx}
\usepackage{algpseudocode}
\usepackage{listings}
\usepackage{float}
\usepackage{array}
\usepackage{xspace}
\usepackage{subcaption}
\usepackage{enumitem}
\usepackage{url}
\usepackage{nicefrac}
\usepackage{microtype}
\usepackage{tabularx}

\usepackage{parskip}
\usepackage{hyperref}
\usepackage[capitalize]{cleveref}

\usepackage[capitalize]{cleveref}
\crefname{section}{Sec.}{Secs.}
\Crefname{section}{Section}{Sections}
\Crefname{table}{Table}{Tables}
\crefname{table}{Tab.}{Tabs.}

\def\eg{\textit{e.g.}\@\xspace} 
\def\ie{\textit{i.e.}\@\xspace} 
\newcommand{\method}{\emph{ SoccerLens}\@\xspace}

%% file: sections/0_abstract.tex
\begin{abstract}
Vision-language models (VLMs) have recently shown strong potential in soccer video understanding.
However, given the high complexity of soccer videos due to large viewpoint variations, rapid shot transitions, and cluttered scenes, it remains unclear on whether VLMs rely on meaningful visual evidence or exploit spurious correlations and shortcut learning.
Existing evaluation protocols focus primarily on classification accuracy and do not assess visual grounding.
To address this limitation, we introduce \method, a benchmark for grounded soccer video understanding.
The benchmark contains annotated video segments spanning $13$ common soccer events, with structured visual cues organized into three levels of semantic relevance.
We further extend the attribution method of Chefer~\cite{Chefer_2021_ICCV} to jointly model spatial and temporal attention, and introduce evaluation metrics that measure whether model attention aligns with annotated cues or drifts toward spurious regions.
Our evaluation of state-of-the-art soccer VLMs shows that, despite strong classification accuracy, current models fail to exceed $50\%$ grounding performance even under the loosest cue definitions and consistently underutilize temporal information.
These results reveal a substantial gap between predictive performance and true visual grounding, highlighting the need for grounded evaluation in complex spatio-temporal domains such as soccer.

\end{abstract}

%% file: sections/1_intro.tex
\section{Introduction}\label{sec:intro}

Vision-based analysis has become a central paradigm for understanding soccer games, as it enables scalable and non-intrusive extraction of information directly from broadcast video. 
Early approaches predominantly relied on tracking-based pipelines \cite{lu2013learning, d_orazio2009ball, kim2004ball}, which localize players, the ball and relevant landmarks to derive higher-level game semantics. 
While effective in controlled settings, these methods require extensive engineering and remain sensitive to variations in camera viewpoints, occlusions, and scene complexity, limiting their robustness in real-world broadcast scenarios. 
These limitations motivate the development of more flexible approaches that can directly reason over raw visual content.

\begin{figure}
    \centering
    \includegraphics[width=\linewidth]{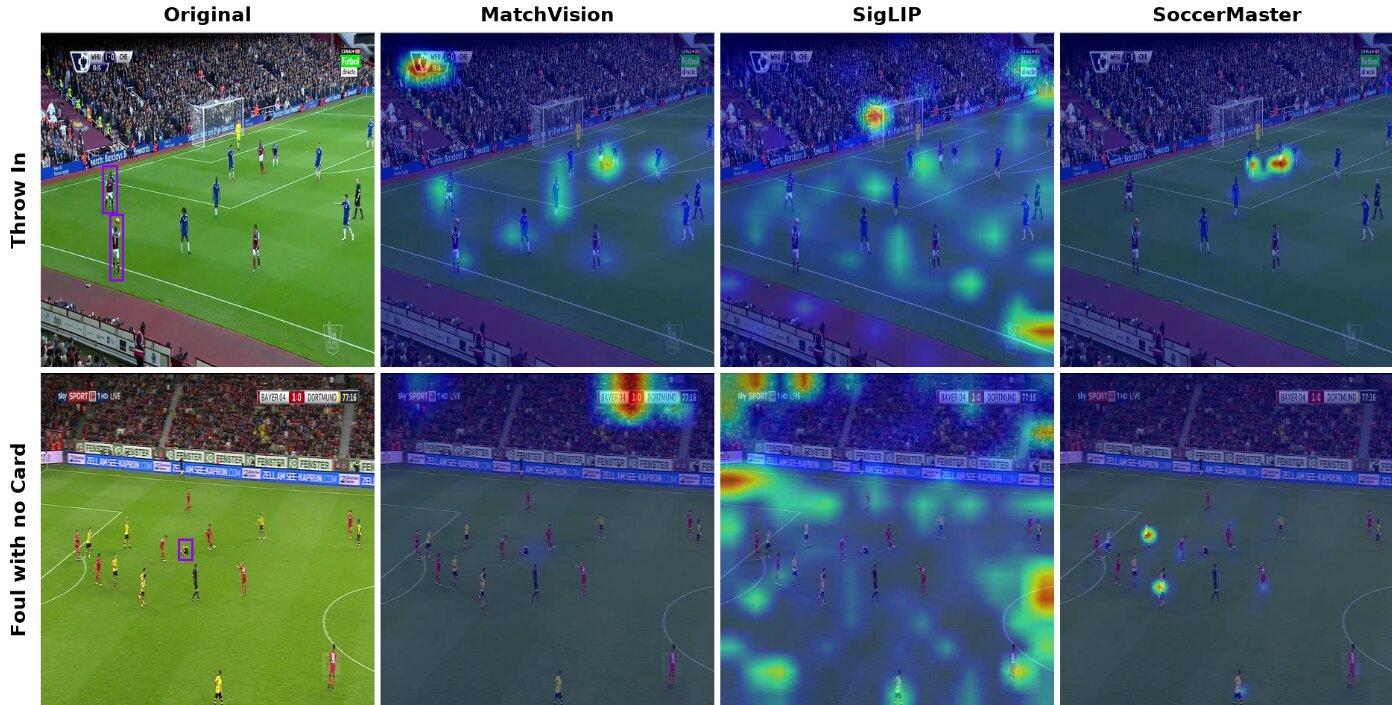}
    \caption{State-of-the-art soccer VLMs such as MatchVision \cite{rao2025universalsoccervideounderstanding} and SoccerMaster \cite{yang2025soccermaster} attend to irrelevant regions when detecting a ``throw-in'' event rather than focusing on the actual player that is performing the action. For the ``Foul with no card'' class, the models failed to identify the offending and injured players and the referee.} 
    
    \label{fig:intro}
\end{figure}

Recent advances in vision-language models (VLMs) \cite{alayrac2022flamingo, radford2021clip, zhai2023siglip, li2022blip} have significantly improved the ability to analyze raw videos under varying conditions. 
Owing to their strong generalization capabilities, VLMs have been increasingly adopted for soccer understanding tasks, including action spotting and event classification \cite{yang2025soccermaster,rao2025universalsoccervideounderstanding}. 
Despite their strong predictive performance, we observe that these models often rely on spurious correlations and attend to irrelevant cues, failing to consistently ground their predictions in semantically meaningful visual evidence.
\Cref{fig:intro} illustrates such a failure case, where state-of-the-art soccer VLMs focus on irrelevant regions when detecting different events (\eg throw-in and foul), instead of attending to players performing or undergoing the action. 
We observed that this behavior is common in far-view shots, where limited visual detail encourages models to exploit shortcuts rather than rely on causal visual cues.

These limitations highlight the need for evaluating not only predictive performance of soccer VLMs but also whether their decisions are grounded in meaningful visual evidence. 
To this end, we introduce \method, a benchmark for grounded soccer video understanding that goes beyond prediction accuracy. 
\method consists of annotated video segments covering 13 common soccer events, where each sample is associated with structured visual cues at three levels: primary cues that are essential for identifying an event, secondary cues that provide supporting context, and common cues that are shared across multiple events. 
For example, in a throw-in, a primary cue corresponds to the individual players performing the action, a secondary cue corresponds to both players together and the ball, and a common cue includes contextual elements such as the out-of-play event and the referee signaling it.
These annotations enable fine-grained evaluation of whether model predictions align with semantically relevant regions in the video.

Given that soccer VLMs incorporate temporal attention to model temporal dynamics, assessing visual grounding requires capturing both spatial and temporal dependencies. 
To this end, we extend the explainability method of Chefer \cite{Chefer_2021_ICCV} to account for the temporal dimension, which we denote as \emph{Chefer-T}. 
Building on this, we introduce evaluation metrics that quantify whether VLMs attend to relevant visual cues across both spatial and temporal dimensions. 
These metrics serve as grounding measures that assess the reliability of VLM predictions and are designed to complement standard predictive accuracy.

We evaluate several state-of-the-art soccer VLMs and find that, despite achieving competitive classification performance, they frequently rely on irrelevant visual cues, indicating the presence of shortcut learning. 
In particular, models fail to exceed $50\%$ grounding even at the loosest level of visual cues (\ie common cues), suggesting that their predictions are not consistently based on relevant evidence. 
Furthermore, our analysis shows that these models underutilize temporal information as suggested by their low scores on our proposed temporal grounding metric.
These findings reveal a gap between predictive performance and true visual understanding, motivating the need for more reliable and interpretable models. 

We expect that \method will serve as a practical benchmark for grounded video understanding, providing a challenging testbed that encourages the development of models that are both accurate and interpretable in complex real-world scenarios.
\footnote{We will release the dataset and evaluation suite to support the development of more reliable models.}

\noindent Our contributions can be summarized as follows:
\begin{enumerate}
    \item We introduce \method, a benchmark for grounded soccer video understanding, consisting of annotated video segments for $13$ common events with structured visual cues at three levels, namely primary, secondary, and common cues. 
    \item We extend the explainability method of Chefer et al. \cite{Chefer_2021_ICCV} to incorporate the temporal dimension, resulting in \emph{Chefer-T}, which enables spatiotemporal attribution for soccer VLMs. 
    \item We propose evaluation metrics that quantify visual grounding by measuring the alignment between model attention and annotated cues across both spatial and temporal dimensions. 
    \item We conduct a comprehensive evaluation of state-of-the-art soccer VLMs and show that, despite strong predictive performance, they exhibit limited grounding, failing to exceed $50\%$ even at the loosest level of cues and underutilizing temporal information. 
\end{enumerate}

%% file: sections/2_related.tex
\section{Related Work}

\subsection{Soccer Video Understanding}
Computer vision in sports has evolved from hand-engineered pipelines to large-scale data-driven benchmarks spanning localization, tracking, calibration, pose estimation, tactical analysis, and event recognition~\cite{thomas2017computer,naik2022comprehensive}.
Early soccer systems relied on task-specific modules, making performance dependent on assumptions about camera setup, broadcast style, and competition context~\cite{d_orazio2009ball,lu2013learning}.
SoccerNet established a standard benchmark for soccer video understanding with $500$ broadcast games and temporally grounded annotations for goals, cards, and substitutions~\cite{giancola2018soccernet}.
Later releases expanded to dense actions, camera segmentation, replay grounding, multi-view spatial annotations, and re-identification~\cite{deliege2021soccernetv2,cioppa2022scaling}.
The SoccerNet challenges further cover temporal, spatial, and identity tasks, including spotting, replay grounding, pitch localization, calibration, and tracking~\cite{giancola2022soccernet}.
Recent extensions introduce group activity recognition, ball action anticipation, and multi-view foul classification~\cite{karki2025pixels,dalal2025action,held2023vars}.
These benchmarks make soccer understanding measurable across multiple dimensions, but evaluation focuses on outputs such as labels or timestamps rather than the evidence supporting them.
As a result, models can succeed by exploiting spurious correlations (\eg broadcast overlays) instead of event-defining cues.
\method introduces a benchmark of visual cues related to different soccer events, enabling evaluation of whether predictions are grounded in event-related evidence or other spurious cues.

\subsection{Soccer Vision-Language Models}

Vision-language models align visual inputs with textual descriptions, enabling classification, retrieval, captioning, and multimodal reasoning after large-scale pretraining~\cite{radford2021clip,alayrac2022flamingo,li2022blip,zhai2023siglip}.
This paradigm is well suited to soccer, where broadcast videos are naturally paired with commentary, annotations, and structured metadata.

Early soccer-specific work leverages language to move beyond fixed event labels toward richer semantic descriptions.
SoccerNet-Caption~\cite{mkhallati2023soccernetcaption} introduced dense captioning by aligning long videos with temporally localized commentary, while GOAL~\cite{Qietal2023} and MatchTime~\cite{rao2024matchtimeautomaticsoccergame} generate commentary grounded in match context.
Language-based interaction is further explored through retrieval and question answering systems such as SoccerRAG~\cite{strand2024soccerrag} and SoccerChat~\cite{gautam2025soccerchat}.
X-VARS~\cite{held2024xvars} is closely related to our setting, as it explains refereeing decisions through multimodal question answering over multi-view foul scenarios.

Recent soccer foundation models learn domain-specific representations that support multiple tasks.
MatchVision~\cite{rao2025universalsoccervideounderstanding} and SoccerMaster~\cite{yang2025soccermaster} combine visual and language supervision for event understanding, commentary generation, and spatial reasoning.
However, evaluation remains focused on final task performance, such as classification accuracy or generation quality, without distinguishing predictions based on event-defining evidence from those based on contextual shortcuts.
For example, a goal may be inferred from broadcast cues such as crowd reactions or broadcast goal panel rather than from the ball crossing the goal line.
\method addresses this limitation by separating event-defining evidence from supporting and contextual cues.

\subsection{Explainability and Visual Grounding}

In vision-language models (VLMs), correct predictions do not necessarily imply correct visual grounding. 
A model may assign the correct label while relying on contextual signals that correlate with the event rather than on the visual evidence that defines it, for example predicting a tennis playing action based on the court instead of the player. 
Such behavior has been observed in early VLMs like CLIP, which often exploit spurious correlations and dataset biases instead of semantically meaningful cues \cite{li2023clip, yuksekgonul2023when, shrestha2020negative, wu2025segdebias}. 
Explainability methods are therefore needed to assess whether predictions are supported by appropriate visual regions.

Attribution methods estimate the contribution of input features to model predictions. 
Integrated Gradients~\cite{sundararajan2017axiomatic} assigns importance through path-integrated gradients, Grad-CAM~\cite{Selvarajuetal2016} produces class-discriminative heatmaps, Score-CAM~\cite{wang2020scorecam} scores activation maps without gradients, and RISE~\cite{petsiuk2018rise} estimates importance via randomized masking. 
For transformer-based models, attention rollout and attention flow~\cite{abnar2020quantifying} approximate information propagation, while Chefer et al.~\cite{Chefer_2021_ICCV} extend relevance propagation to multimodal architectures.

In soccer videos, relevant evidence is event-dependent and often entangled with contextual shortcuts such as celebrations, broadcast graphics, or editing patterns. 
Generic explainability methods indicate where a model attends, but not whether these regions are event-defining, supportive, or merely correlated. 
\method addresses this limitation with cue-level annotations that distinguish primary, secondary, and common cues, enabling quantitative evaluation of shortcut behavior. 
Furthermore, since soccer VLMs incorporate temporal attention, we extend Chefer et al.~\cite{Chefer_2021_ICCV} to capture temporal dependencies, denoted as \emph{Chefer-T}.

%% file: sections/3_dataset.tex
\section{\method Benchmark} \label{sec:benchmark}

In soccer event recognition, evaluation is typically conducted using task-specific metrics such as Top-\textit{k} accuracy, as in SoccerNet \cite{giancola2018soccernet}. 
However, such metrics do not assess whether predictions are grounded in the visual evidence of the video. 
Consequently, a model may correctly classify a \textit{Goal} event based on contextual cues such as player celebrations or audience reactions, while failing to capture the decisive moment of the ball crossing the goal line. 
To develop more reliable models, it is therefore essential to evaluate whether predictions are grounded both spatially and temporally. 
To fill this gap, we introduce \method, the first dataset that augments existing soccer datasets (\eg MatchTime) with bounding box annotations for structured visual cues associated with $13$ common soccer events. 
In addition, we define a set of evaluation metrics that quantify the spatial and temporal grounding performance of VLMs alongside predictive accuracy. 
Together, the dataset and metrics establish a benchmark for the systematic evaluation of the spatio-temporal grounding capabilities of soccer Vision-Language Models (VLMs)
\footnote{\method dataset is available at \url{https://github.com/IsmaelElsharkawi/SoccerLensDataset}}.

\subsection{Building the Dataset}
We construct our dataset based on MatchTime \cite{rao2024matchtimeautomaticsoccergame}, which contains videos of diverse soccer events across multiple leagues. 
Following the MatchTime protocol, we use a clip duration of $30$ seconds with a sampling rate of $1$ frame per second, resulting in $30$ frames per clip. 
From this source, we select $200$ clips spanning $13$ common soccer events, including \texttt{corner}, \texttt{goal}, \texttt{yellow card}, \texttt{red card}, \texttt{foul (no card)}, \texttt{lead to corner}, \texttt{free kick}, \texttt{substitution}, \texttt{injury}, \texttt{foul leading to penalty}, \texttt{throw-in}, \texttt{penalty}, and \texttt{second yellow card}. 
Each frame is annotated by professional annotators with three categories of visual cues:

\begin{itemize}
    \item \textbf{Primary Cues (P-Cues)} encapsulate the most salient visual features relevant to an event in a highly localized manner (\eg the player performing throw-in)
    \item \textbf{Secondary Cues (S-Cues)} are less localized than primary cues. In addition, if there are multiple primary cues in a frame, these should be contained within a secondary cue (\eg an encapsulation of player throwing and receiving the ball)
    \item \textbf{Common Cues (C-Cues)} are visual features that can be temporally associated with an event; but do not represent the event itself. It is important to label these separately to be able to identify a VLM's power to separate the event from distractions. (\eg the referee reacting to the ball leaving the field before the throw in)
\end{itemize}

\begin{figure}[t]
    \centering
    \includegraphics[width=\linewidth]{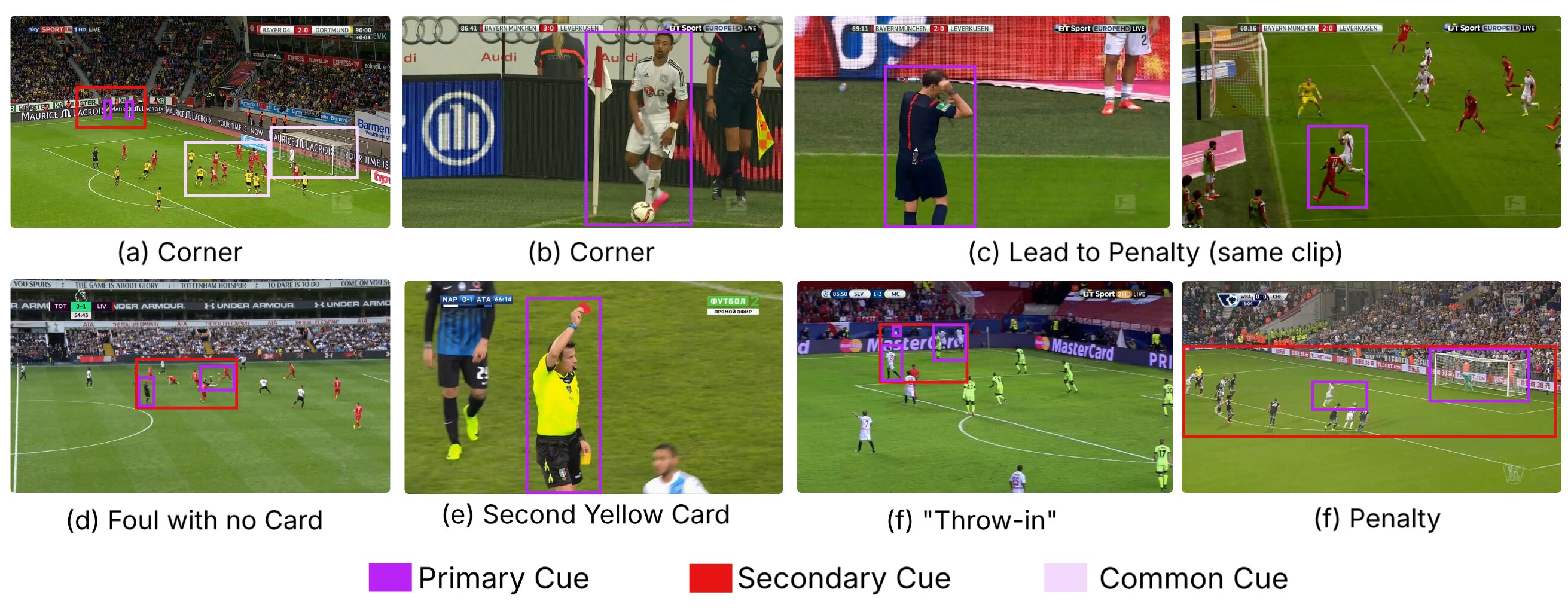}
    \caption{There are three types of annotations: Primary Cue, Secondary Cue and Common Cue. This figures illustrates examples for each of the three types.}
    \label{fig:cues_examples}
\end{figure}

\Cref{fig:cues_examples} shows some examples of different cues for several events.
In total, the dataset contains $2{,}209$ annotated frames\footnote{Frames with no cues are skipped.}. 
The distribution of clips per event, along with detailed definitions of primary, secondary, and common cues, across different events is provided in Table \ref{tab:cues-table}.

\subsection{Evaluation Metrics}
\label{sec:evaluation_criteria}

Having defined the visual cues, we next introduce evaluation metrics that quantify whether a model attends to these cues both spatially and temporally when making predictions. 
These metrics are model-agnostic and can be computed using any explainability mechanism, including raw attention maps, attribution methods such as Chefer \cite{Chefer_2021_ICCV}. 

\textbf{1. Energy inside bounding box} measures the proportion of attribution mass that falls within the annotated regions. 
\begin{equation}
    \text{Energy (\%)} = \dfrac{\sum_{i,j} E_{i,j}\cdot B_{i,j}}{\sum_{i,j} E_{i,j}}
\end{equation}
where $E_{i,j}$ denotes the relevancy map intensity at pixel $(i,j)$ and $B_{i,j}$ is a binary indicator specifying whether the pixel lies inside a ground-truth bounding box. 

\textbf{2. Pointing} evaluates whether the most salient pixel identified by the model aligns with the annotated regions. 
For each frame, it is defined as a binary indicator that takes the value $1$ if the pixel with maximum attribution lies within any bounding box, and $0$ otherwise. 

\textbf{3. Spatial Intersection over Union (S-IoU)} quantifies the overlap between predicted salient regions and ground-truth bounding boxes. 
\begin{equation}
    \text{S-IoU} = \frac{| P_p \cap P_{gt} |}{| P_p \cup P_{gt} |}
\end{equation}
where $P_p$ denotes the set of predicted salient pixels and $P_{gt}$ denotes the set of pixels within the ground-truth bounding boxes. 
The predicted set $P_p$ is obtained by thresholding the relevancy map at $50$\% of its maximum value within each frame. 

\begin{figure}[t]
    \centering
    \includegraphics[width=0.5\linewidth]{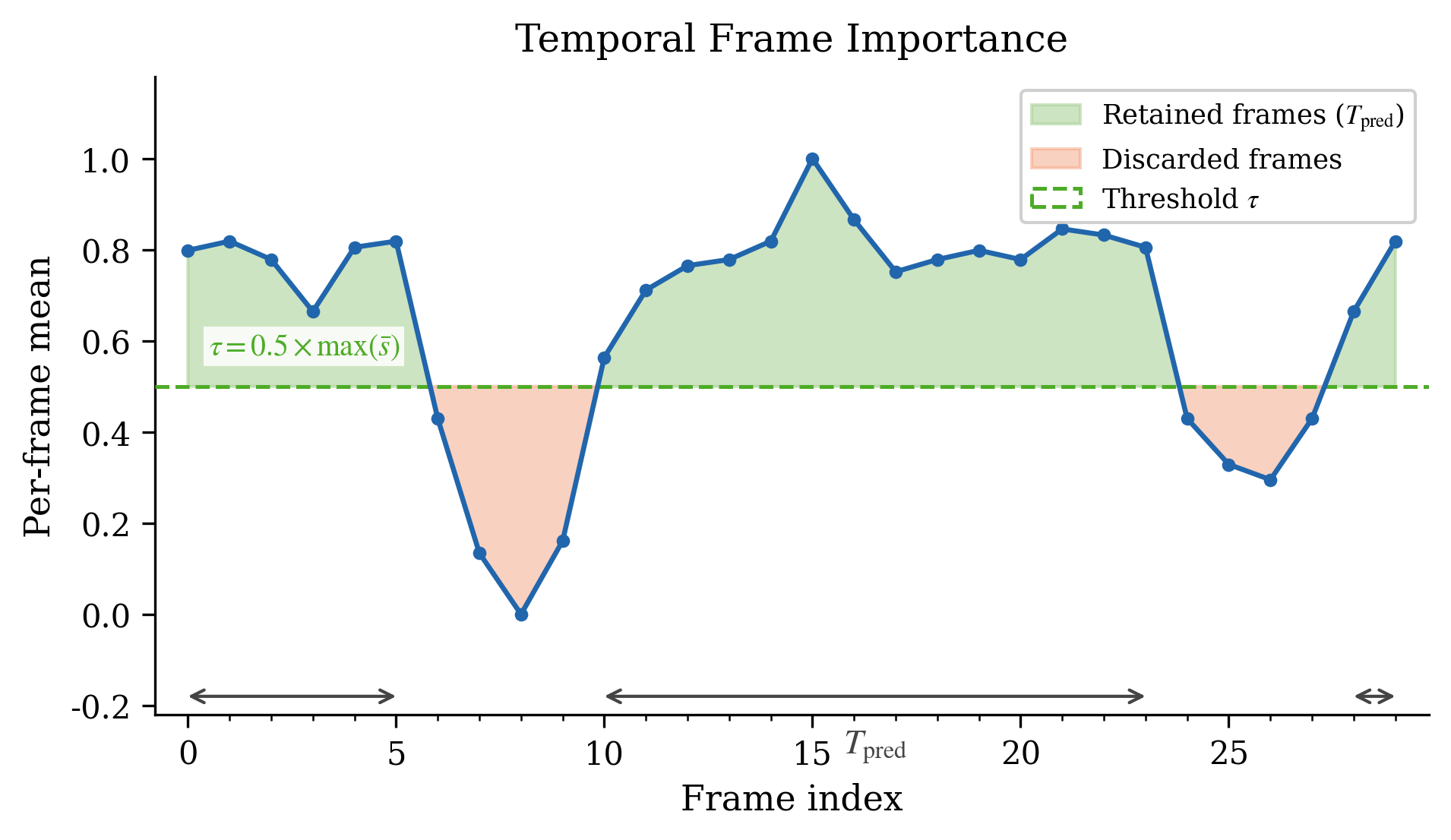}
    \caption{$T_{pred}$ frames are the frames with the average energy greater than $0.5\times max(\bar{s})$, where $\bar{s}$ is the per-frame mean.}
    \label{fig:tiou}
\end{figure}

\textbf{4. Temporal Intersection over Union (T-IoU)} measures the alignment between frames deemed salient by the model and frames containing annotated cues. 
\begin{equation}
    \text{T-IoU} = \frac{| T_{gt} \cap T_{pred} |}{| T_{gt} \cup T_{pred} |}
\end{equation}
where $T_{gt}$ is the set of frames containing at least one bounding box, and $T_{pred}$ is the set of frames whose average attribution exceeds $50$\% of the maximum frame-level mean attribution within the clip.
We use this adaptive threshold as different models will have different levels of attribution.
An illustration of the temporal frame importance is shown in \Cref{fig:tiou}.

%% file: sections/4_prel.tex
\section{Explainability for Soccer VLMs}

Explainability remains an open research problem aimed at understanding how models arrive at their predictions.
While our benchmark is compatible with a wide range of explainability methods, we focus on attribution-based approaches, specifically the method of Chefer et al.~\cite{Chefer_2021_ICCV}, which has been shown to be effective for transformer-based architectures.
However, the original Chefer formulation only accounts for spatial attention and therefore cannot capture temporal dependencies that are essential for modeling dynamic events in soccer videos.
To address this limitation, we extend this method to incorporate temporal attention, resulting in our proposed variant, \emph{Chefer-T}.

\subsection{Transformer-MM-Explainability (Chefer et al.)}
\label{subsec:chefer}
A $30$-second input video is represented as $x \in \mathbb{R}^{3\times H\times W\times T}$, where $T=30$. We adopt the gradient-weighted relevance propagation method of Chefer, applied independently to each of the $T$ frames of $x$.
For a given frame $t$ and layer $l$, let $A_l^{(t)} \in \mathbb{R}^{K \times N \times N}$ be the multi-head attention map, where $K$ is the number of attention heads and $N$ is the number of tokens, and let $\partial f / \partial A_l^{(t)}$ be its gradient with respect to the target-class logit $f$. The head-averaged positive contribution is
\begin{equation}
\bar{A}_l^{(t)} \;=\; \mathbb{E}_h\!\left[\mathrm{ReLU}\!\Bigl(\dfrac{\partial f}{\partial A_l^{(t,h)}} \odot A_l^{(t,h)}\Bigr)\right] \;\in\; \mathbb{R}^{N\times N},
\label{eq:chefer_cam}
\end{equation}
where $\odot$ denotes elementwise multiplication and $\mathbb{E}_h$ denotes the expectation across the $K$ attention heads. Intuitively, $\bar{A}_l^{(t)}$ keeps only the attention edges whose direction agrees with the gradient of the prediction, so each entry $\bar{A}_l^{(t)}[i, j]$ is large when input token $j$ both attends to output token $i$ and contributes positively to the target logit.
A per-frame token-to-token relevance matrix $R_t^{\text{spat}} \in \mathbb{R}^{N \times N}$ is initialized as the identity matrix and iteratively refined across layers using $R_t^{\text{spat}} \leftarrow R_t^{\text{spat}} + \bar{A}_l^{(t)} R_t^{\text{spat}}$.
In a standard ViT, the row $R_t^{\text{spat}}[0,\,1\!:\!N]$, reshaped into a 2D patch grid, yields the per-frame spatial relevance.

\subsection{Adapting Chefer's for Temporal Attention (Chefer-T)}

\label{subsec:chefer_adaptations}
To extend Chefer to factored space-time soccer transformers, we introduce three key modifications: a revised extraction mechanism at the head, temporal relevance propagation at each spatial position, and a two-stream combination strategy.

\paragraph{Pooling-head extraction.}
The SigLIP-family encoders used in our benchmark do not include a CLS token.
Instead, we extract relevance using the pooling head.
Let $A_{\text{pool}}^{(t)} \in \mathbb{R}^{K \times 1 \times N}$ denote the attention from the learnable probe to the $N$ patch tokens, and let $\bar{A}_{\text{pool}}^{(t)} \in \mathbb{R}^{1 \times N}$ denote its head-averaged positive contribution computed via Equation~\ref{eq:chefer_cam}.
The resulting per-frame spatial relevancy map is given by
\begin{equation}
\mathcal{H}_t^{\text{spat}} \;=\; \bar{A}_{\text{pool}}^{(t)}\, R_t^{\text{spat}} \;\in\; \mathbb{R}^{1 \times N}.
\label{eq:chefer_pool}
\end{equation}

\paragraph{Per-position temporal propagation.}
Both MatchVision and SoccerMaster include blocks that factor self-attention into temporal and spatial components.
In these architectures, temporal attention operates across the $T$ frames at each fixed spatial position, which is critical for capturing event dynamics such as passes, shots, and transitions.
To capture temporal relevance, we follow~\cite{ISTVT} and maintain a per-position temporal relevance matrix $R_p^{\text{temp}} \in \mathbb{R}^{T \times T}$ for each spatial index $p \in \{1, \ldots, N\}$, initialized as the identity.
Each factored block updates this matrix as $R_p^{\text{temp}} \leftarrow R_p^{\text{temp}} + \bar{A}_{l,p}^{\text{temp}} R_p^{\text{temp}}$, where $\bar{A}_{l,p}^{\text{temp}}$ is computed using Equation~\ref{eq:chefer_cam} on the corresponding $T \times T$ attention slice.
The classifier head applies temporal self-attention over pooled frame embeddings, producing contributions $\bar{A}_l^{\text{head}} \in \mathbb{R}^{T \times T}$ that are broadcast across all spatial positions using the same update rule.

\paragraph{Two-stream combination.}
The final per-frame attribution map is obtained by combining spatial relevance with temporal weights derived from $\{R_p^{\text{temp}}\}$.
Specifically, we compute
\begin{equation}
w_t \;=\; \frac{1}{N\,T} \sum_{p=1}^{N} \sum_{t'=1}^{T} R_p^{\text{temp}}[t',\,t],
\qquad
\hat{\mathcal{H}}_t \;=\; \mathrm{MinMax}\!\bigl(\mathcal{H}_t^{\text{spat}}\bigr) \cdot \frac{w_t}{\max_{t'} w_{t'}},
\label{eq:chefer_combine}
\end{equation}
Reading the first formula: $R_p^{\text{temp}}[t',\,t]$ measures how strongly source frame $t$ contributes to query frame $t'$ at spatial position $p$; averaging over both $p$ and $t'$ collapses these per-position, per-query contributions into a single scalar that quantifies the total temporal importance of frame $t$ for the prediction. The second formula then renormalizes this scalar to lie in $[0,1]$ before scaling the spatial relevance, so the brightest frame in the final attribution corresponds to the most temporally important one.
Min-max normalization is applied prior to temporal scaling to preserve temporal variation, as applying normalization afterward would suppress the temporal signal.
Finally, $\hat{\mathcal{H}}_t$ is reshaped into a 2D patch grid and upsampled to resolution $H \times W$ using bilinear interpolation, yielding $\hat{\mathcal{H}}_t \in \mathbb{R}^{H \times W}$. The full procedure is given as Algorithm~\ref{alg:chefer_t} in Appendix~\ref{app:chefer_t_algorithm}.

%% file: sections/6_exp.tex
\section{Experiments}
\label{sec:experiments}
\subsection{Experimental Setup}
We evaluate state-of-the-art soccer VLMs on our \method benchmark, namely MatchVision \cite{rao2025universalsoccervideounderstanding} and SoccerMaster \cite{yang2025soccermaster}.
We also evaluate SigLIP \cite{zhai2023siglip} as a standard non-specialized VLM.
We provide more details on these models in supplementary \Cref{sec:models}. Our code is available at \url{https://github.com/IsmaelElsharkawi/SoccerExplainability}.

\subsection{Quantitative Results}
We report the scores of our proposed metrics on Chefer-T in Table \ref{tab:horizontal_space_saver}. There is a comparison of the standard Chefer and Chefer-T performance presented in Table \ref{tab:chefer_spatial_vs_temporal}.

\paragraph{Spatial Grounding.}
MatchVision achieves the highest Energy and S-IoU scores, indicating that it has a higher concentration of attention at different cues (Table \ref{tab:horizontal_space_saver}). However, this score does not exceed $39\%$ when including all visual cues, indicating a huge gap and attention to irrelevant cues.
Whereas SoccerMaster achieves the highest pointing accuracy indicating that it has a higher coverage of attention, but with less density.
These results highlight that despite SoccerMaster having higher prediction accuracy than MatchVision, its spatial grounding capability is less.
This inversion suggests a systematic decoupling of accuracy and grounding under domain-specialized training.

\paragraph{Temporal Grounding.}
SigLIP, which has no temporal attention, achieves the best T-IoU by a large margin. We argue this reveals a concrete failure of task-specialized training: MatchVision and SoccerMaster over-attend to event-correlated frames bearing broadcast shortcuts, misaligning with the true event window. Table \ref{tab:chefer_spatial_vs_temporal} confirms this; Chefer-T yields only marginal T-IoU gains over standard Chefer, indicating temporal attention carries little grounding signal.

\begin{table}[!t]
    \centering
    \hspace*{-1.5cm}
    \scriptsize 
    \setlength{\tabcolsep}{1.1pt} 
    
    \begin{minipage}[t]{0.56\textwidth} 
        \centering
        \caption{Chefer-T results for \textit{SoccerLens} benchmark.}
        \label{tab:horizontal_space_saver}
        \begin{tabular}{l ccc ccc ccc}
            \toprule
            & \multicolumn{3}{c}{\textbf{SigLIP}$^\dagger$} & \multicolumn{3}{c}{\textbf{MatchVision}} & \multicolumn{3}{c}{\textbf{SoccerMaster}} \\
            \cmidrule(lr){2-4} \cmidrule(lr){5-7} \cmidrule(lr){8-10}
            \textbf{Metric (\%)} & \textbf{P} & \textbf{P+S} & \textbf{P+S+C} & \textbf{P} & \textbf{P+S} & \textbf{P+S+C} & \textbf{P} & \textbf{P+S} & \textbf{P+S+C} \\
            \midrule
            Energy      & 16.54 & 22.09 & 22.72 & 26.25 & 34.67 & \textbf{39.17} & 24.41 & 34.43 & 37.00 \\
            Pointing    & 24.67 & 30.58 & 28.01 & 30.31 & 39.07 & 44.26 & 35.57 & 47.22 & \textbf{49.77} \\
            S-IoU       & 3.39  & 2.81  & 2.94  & 6.81  & 5.87  & \textbf{7.41}  & 3.95  & 3.61  & 3.49  \\
            T-IoU       & 19.27 & 19.34 & \textbf{28.67} & 11.85 & 11.88 & 15.33 & 8.08  & 8.10  & 15.36 \\
            \midrule
            Acc. (\%) & \multicolumn{3}{c}{7.5} & \multicolumn{3}{c}{62.5} & \multicolumn{3}{c}{66.0} \\
            \bottomrule
        \end{tabular}
        \par\vspace{2pt}
        \raggedright \tiny \textit{Note: \textbf{P}: Primary, \textbf{S}: Secondary, \textbf{C}: Common Cues. $\dagger$ SigLIP has no temporal attention and reported results are                standard Chefer.}
    \end{minipage}
    \hspace{1em} 
    \begin{minipage}[t]{0.38\textwidth} 
        \centering
        \vspace{2.5em}
        \caption{T-IoU: Standard Chefer vs Chefer-T.}
        \label{tab:chefer_spatial_vs_temporal}
        \begin{tabular}{l ccc ccc}
            \toprule
            & \multicolumn{3}{c}{\textbf{Chefer}} & \multicolumn{3}{c}{\textbf{Chefer-T}} \\
            \cmidrule(lr){2-4} \cmidrule(lr){5-7}
            \textbf{Model} & \textbf{P} & \textbf{P+S} & \textbf{P+S+C} & \textbf{P} & \textbf{P+S} & \textbf{P+S+C} \\
            \midrule
            \textbf{MatchVision}  & 11.56 & 11.59 & 14.98 & 11.85 & 11.88 & \textbf{15.33} \\
            \textbf{SoccerMaster}  & 7.89  & 7.90  & 15.11 & 8.08  & 8.10  & \textbf{15.36} \\
            \bottomrule
        \end{tabular}
    \end{minipage}
\end{table}

\subsection{Qualitative Analysis}
\label{sec:qualitative}
\begin{figure}[!t]
    \centering
    \includegraphics[width=0.9\linewidth]{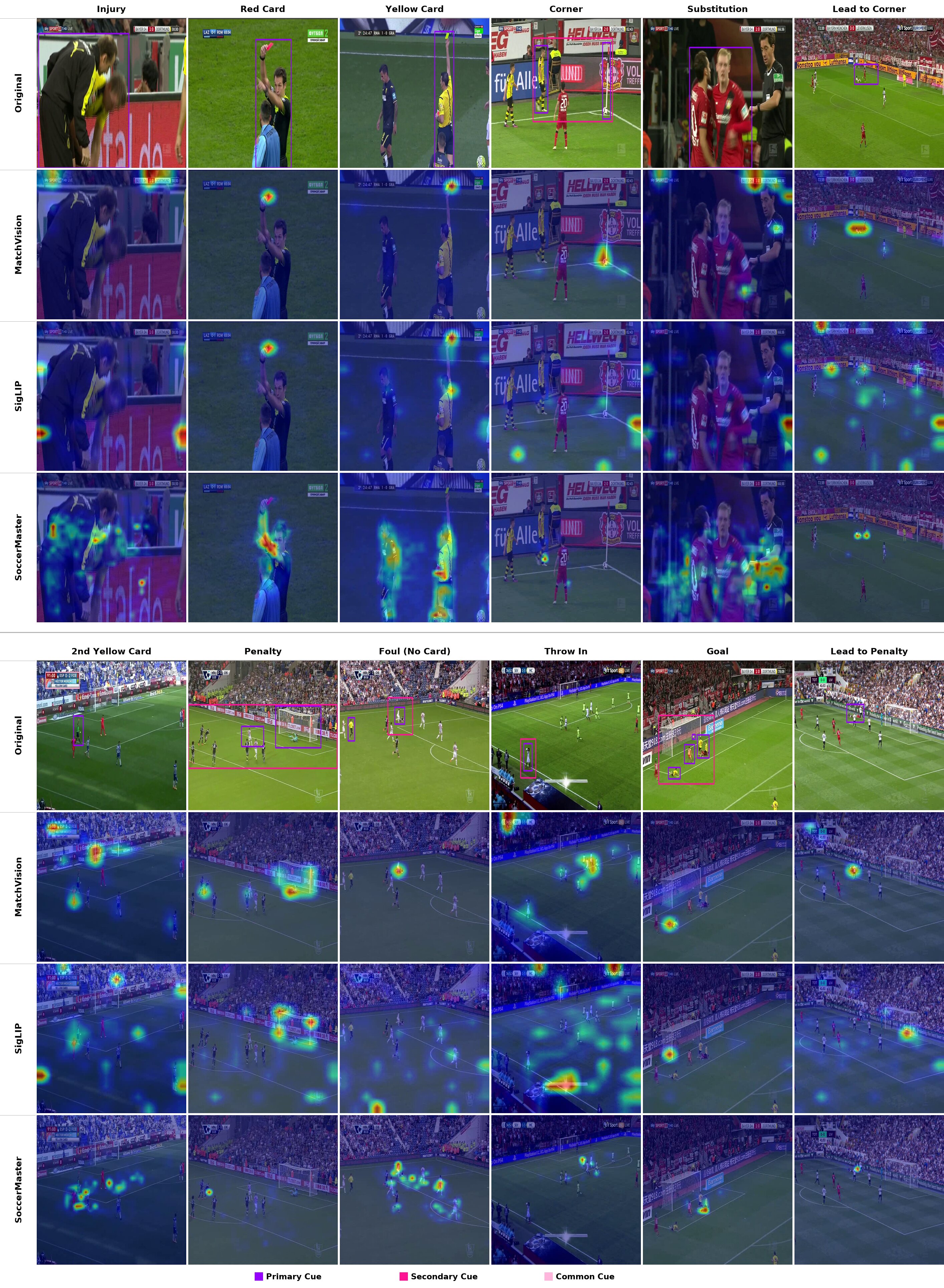}
    \caption{Relevancy map overlays for different classes.}
    \label{fig:allClasseseVisualization}
\end{figure}

A model should \textit{focus} on primary and secondary visual cues, while only \textit{observing} common visual cues. Figures \ref{fig:allClasseseVisualization} show the Chefer-T relevancy map overlays for one randomly selected example per-class.

\paragraph{Success Cases}
Certain classes are more easily recognizable than others.
For example, SoccerMaster correctly identifies the Injury class and attends to the medical staff members, as shown in the top first column of \Cref{fig:allClasseseVisualization}.
Similarly, the Red Card and Yellow Card classes are consistently recognized across all models, with attention focused on the card held by the referee.

\paragraph{Failure Cases}
In several failure cases, models ignore primary cues in favor of less relevant regions. For Substitution, neither player in the exchange is recognized by any model. For Penalty, SoccerMaster attends to a running player rather than the shooter. For Throw-in, MatchVision and SigLIP are distracted by the TV panel and irrelevant players, consistent with Figure 2. The Goal and Corner classes notable exceptions: MatchVision and SoccerMaster attend to different yet semantically valid primary cues. For Corner, SoccerMaster identifies only the player's foot while MatchVision focuses solely on the corner flag, where ideally both models should attend to both cues. For Goal, MatchVision focuses on the assist and SoccerMaster on the scorer.

\paragraph{Camera-Angle Variations}
Primary visual cues that require fine-grained spatial localization are more often difficult to detect. For example, in the Lead to Corner class on the top sixth column (Figure \ref{fig:allClasseseVisualization}), the ball leaving the pitch is only identified by MatchVision and SoccerMaster, while being missed by SigLIP. Similarly, the Red Card held by the referee for the Second Yellow Card is not highlighted in any relevancy map, because the camera angle is far away from the referee. The model should highlight both yellow and red cards held by the referee, which are missed by all models.

%% file: sections/7_discussion.tex
\section{Discussion}\label{sec:discussion}
Across both quantitative and qualitative results, accuracy and grounding are consistently dissociable. SoccerMaster's higher accuracy but weaker spatial grounding than MatchVision suggests spatial pretraining encourages broad coverage at the expense of event-specific localization. The temporal inversion is more revealing: SigLIP outperforms specialized models on T-IoU despite having no temporal attention, indicating that temporal modules learn to weight label-correlated broadcast frames rather than the event itself, corroborated by the negligible Chefer-T gain for both MatchVision and SoccerMaster in Table \ref{tab:chefer_spatial_vs_temporal}. At the event level, close-up classes (in Figure \ref{fig:allClasseseVisualization}) such as Corner, Yellow Card and Red Card ground better in some cases than far-view classes such as Throw-in and Second Red Card, where attention drifts toward irrelevant players. The consistency of these failure modes across three architecturally distinct models rules out model-specific explanations, pointing instead to a fundamental limitation of training on event labels alone: models exploit any correlating feature regardless of causal relevance, and grounding will only improve if explicitly optimized for.

%% file: sections/8_conclusion.tex
\section{Conclusion}\label{sec6}
We present \method, a benchmark for spatiotemporal visual grounding in soccer VLMs across three hierarchical cue tiers. 
Experiments across SigLIP, MatchVision, and SoccerMaster yield three consistent findings: 1- The accuracy and grounding are dissociable, with the highest-accuracy model showing weaker spatial grounding. 2- The task-specialized training degrades temporal grounding, with a non-temporal model outperforming specialized ones on T-IoU. 3- No model exceeds 50\% grounding at any cue level, across all metrics. 
The consistency of these results across architecturally distinct models rules out model-specific explanations and points to fundamental limitations in how current soccer VLMs reason visually. We hope \method motivates training objectives that explicitly optimize for grounding alongside accuracy.

%% file: sections/appendixA.tex
\newpage
\appendix
\section{Labeling Criteria}
This Appendix presents the criteria used to label the three tiers of visual cues described in Section \ref{sec:benchmark} in Table \ref{tab:cues-table}.

\begin{table}[H]
\caption{Labeling Criteria for small bounding boxes, large bounding boxes and correlated visual cues bounding boxes}
\label{tab:cues-table}
\footnotesize
\begin{tabularx}{\textwidth}{@{} l X X X r @{}}

\toprule
Class & Primary Cues  & Secondary Cues &  Common Cues & Count\\
\midrule
Corner    & Corner flag, player heading to the corner flag, corner play& Larger box around the corner arc area & Players grouped close to net, net or goalkeeper, TV Panel  & 23  \\
Goal    & Shot on goal, ball, assist & Goalmouth area, goalkeeper, net  & Player celebration, scoreboard update, referee pointing to center circle, fans cheering & 24  \\
Yellow Card    & Referee raising the yellow card, player receiving the yellow card & Player(s) involved in the foul and referee grouped together & Foul leading to card, TV panel, player(s) hurt & 20 \\
Foul With No Card  & Foul action, referee & Player(s) involved in foul and the referee& Injured player, free kick, another irrelevant foul within 30-second window & 21 \\
Lead to Corner & Final touch before ball crosses the line and the ball & Player and ball as ball leaves the field & Corner play, referee & 19\\
Free Kick & Kicker, ball, referee, defensive wall of players & Kicker and defensive wall and ball together & Foul leading to free kick & 18\\
Substitution & 4th official's board, Players entering or leaving field & Both players involved in the substitution &  TV panel & 17\\
Injury & Medical staff, referee, player injured & Injured player and medical staff & TV panel, players around the injured player & 16 \\
Foul Lead To Penalty & Foul action, player(s) on ground, referee & Players involved in foul and the referee & Penalty play & 13\\
Penalty & Shot on goal and goalmouth area & Goalmouth area, net and the kicker altogether & Referee, Foul leading to penalty & 12  \\
Throw-in & Player throwing the ball and the player receiving the ball& Both players throwing and receiving the ball & Out-of-play action, referee & 7 \\
Second Yellow Card & Referee raising the cards, referee pointing to tunnel & Player(s) involved in the foul and referee grouped together & Foul leading to card, TV panel, Player receiving card & 7 \\
Red Card & Referee raising the red card, referee pointing to tunnel &Player(s) involved in the foul and referee grouped together & Foul leading to red card, TV panel, Player receiving card & 3\\
\midrule
Total & & & & 200 \\
\bottomrule

\end{tabularx}

\end{table}

%% file: sections/appendixB.tex
\newpage
\section{Relevancy Map Visualizations}

\begin{figure}[H]
    \centering
    \includegraphics[width=\linewidth]{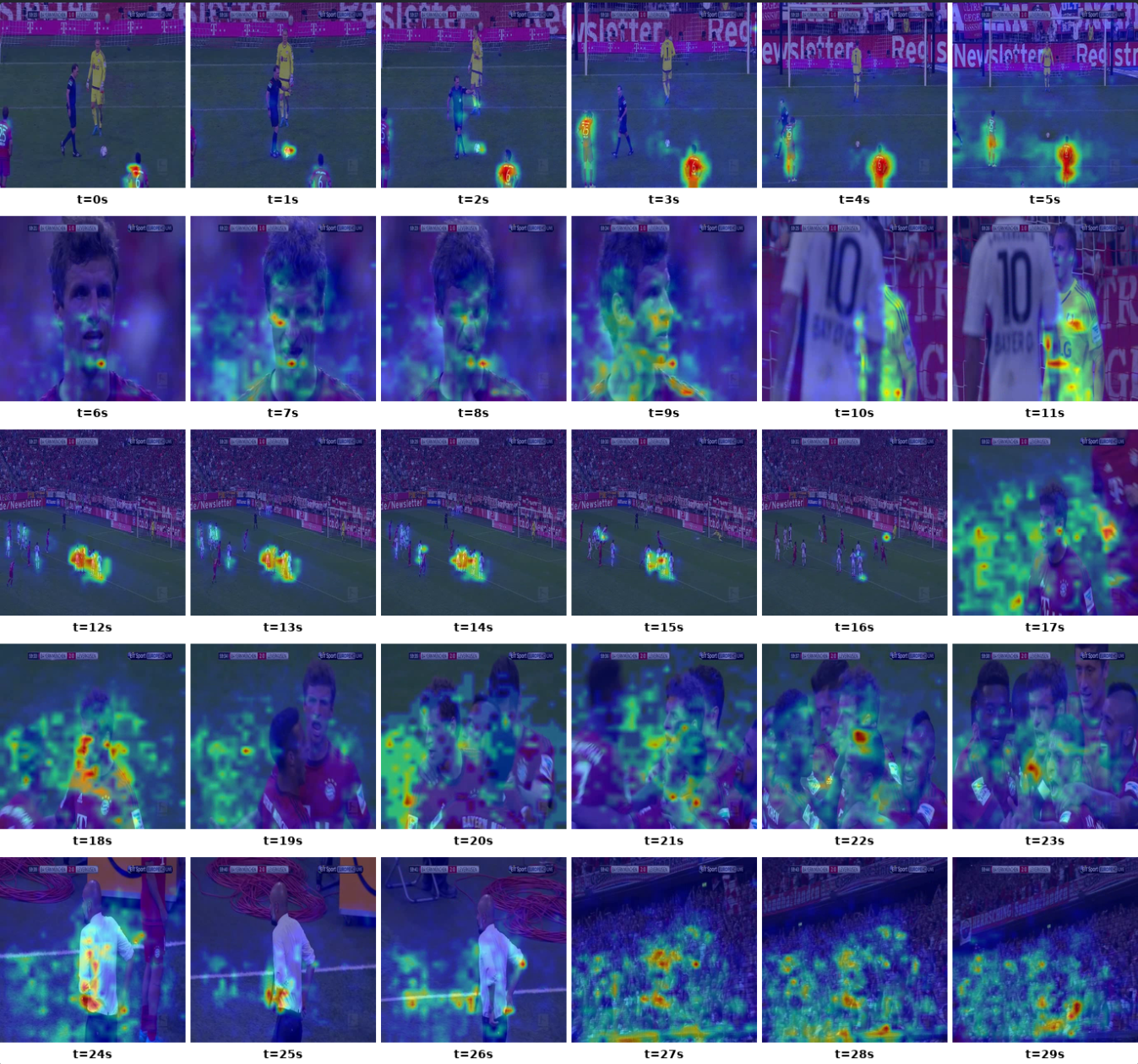}
    \caption{Relevancy map overlays for the 30 seconds of a Goal class using Chefer explainability on SoccerMaster}
    \label{fig:temporal_figure}
\end{figure}

This Appendix presents Figure \ref{fig:temporal_figure}, which shows an example clip for a Goal class, where the Chefer relevancy map overlays are displayed for the 30 sampled frames of the input video. 

\section{Chefer-T Algorithm}
\label{app:chefer_t_algorithm}
Algorithm~\ref{alg:chefer_t} summarizes the full Chefer-T procedure.

\begin{algorithm}[t]
\caption{Chefer-T: per-frame attribution for factored space-time transformers}
\label{alg:chefer_t}
\begin{algorithmic}[1]
\Require Trained model $f$, input video $x$, target class $c$.
\Ensure Per-frame attribution maps $\{\hat{\mathcal{H}}_t\}_{t=1}^{T}$.
\Statex
\State Compute $f(x)$ via forward propagation.
\State Backpropagate from the target-class logit; save attention maps $A_l$ and gradients $\partial f / \partial A_l$ for every backbone and classifier-head block.
\Statex \textit{// Per-frame spatial relevance}
\For{$t = 1$ to $T$}
    \State Initialize $R_t^{\text{spat}} \gets I$.
    \For{each spatial backbone block $l$ in forward order}
        \State Compute $\bar{A}_l^{(t)}$ by Eq.~\ref{eq:chefer_cam}.
        \State $R_t^{\text{spat}} \gets R_t^{\text{spat}} + \bar{A}_l^{(t)} R_t^{\text{spat}}$.
    \EndFor
    \State Compute $\mathcal{H}_t^{\text{spat}}$ by Eq.~\ref{eq:chefer_pool}.
\EndFor
\Statex \textit{// Per-position temporal relevance}
\State Initialize $R_p^{\text{temp}} \gets I$ for each $p \in \{1, \ldots, N\}$.
\For{each factored backbone block $l$}
    \For{$p = 1$ to $N$}
        \State Compute $\bar{A}_{l,p}^{\text{temp}}$ by Eq.~\ref{eq:chefer_cam} on the $T \times T$ attention slice at position $p$.
        \State $R_p^{\text{temp}} \gets R_p^{\text{temp}} + \bar{A}_{l,p}^{\text{temp}}\, R_p^{\text{temp}}$.
    \EndFor
\EndFor
\For{each classifier-head temporal-attention block $l$}
    \State Compute $\bar{A}_l^{\text{head}}$ by Eq.~\ref{eq:chefer_cam}.
    \State Update $R_p^{\text{temp}} \gets R_p^{\text{temp}} + \bar{A}_l^{\text{head}}\, R_p^{\text{temp}}$ for each $p$.
\EndFor
\Statex \textit{// Two-stream combination}
\For{$t = 1$ to $T$}
    \State Compute $w_t$ and $\hat{\mathcal{H}}_t$ by Eq.~\ref{eq:chefer_combine}.
    \State Reshape $\hat{\mathcal{H}}_t$ to a 2D patch grid and upsample to $H \times W$ by bilinear interpolation.
\EndFor
\State \Return $\{\hat{\mathcal{H}}_t\}_{t=1}^{T}$.
\end{algorithmic}
\end{algorithm}

\section{Details on Evaluated Models}
\label{sec:models}
We use SigLIP \cite{zhai2023siglip} as a baseline, whereas MatchVision \cite{rao2025universalsoccervideounderstanding} and SoccerMaster \cite{yang2025soccermaster} are the State-of-The-Art Soccer event classification models. MatchVision \cite{rao2025universalsoccervideounderstanding} uses alternating spatial and temporal attention layers, such that the model is pretrained on event recognition and vision-language alignment using the live commentary, then fine-tuned for event recognition, foul classification and commentary generation. We use the pretrained checkpoint for event recognition. 
Unlike MatchVision which has an input resolution of $224\times 224$ input images, SoccerMaster \cite{yang2025soccermaster} has $512\times 512$. SoccerMaster is pretrained on both Spatial Tasks, namely Athlete Detection and Pitch Registration, and Semantic Tasks, which are vision-language alignment and event classification. The main difference in model architecture is the presence of multiple subsequent spatial attention layers before the alternation of spatial and temporal attention. 
In the case of SigLIP, we use the pretrained checkpoint \textit{google/siglip-base-patch16-224} such that the similarity scores between the text encoding of the class labels and the visual encoding of the 30 input frames are used to get the winning class.

\section{Compute Required}
In our experiments, we use a single NVIDIA RTX Pro 6000 (Google Colab G4 instance) for almost 3 hours for inference only because the evaluation is training-free. We use 30-second clips with 1fps sampling rate for all models.

\section{Limitations }
SoccerLens is constructed from a single broadcast source (MatchTime), and grounding behavior may vary across leagues, production styles, and camera setups. Evaluation is restricted to attribution-based explainability methods; however, the proposed metrics are method-agnostic and directly applicable to other attribution approaches such as RISE or attention rollout. Future work should expand the benchmark to additional broadcast sources and evaluate a broader set of Soccer VLMs and explainability methods.